\documentclass[preprint,12pt,authoryear]{elsarticle}
\usepackage{booktabs}
\usepackage{graphicx}
\usepackage{subcaption}
\usepackage{tabularx}
\usepackage{xcolor}

\usepackage{amssymb}

\journal{Forensic Science International: Digital Investigation}

\begin{document}

\begin{frontmatter}



\title{GenAI Mirage: The Impostor Bias and the Deepfake Detection Challenge in the Era of Artificial Illusions}


\author[inst1,inst2]{Mirko Casu\corref{cor1}}
\author[inst1]{Luca Guarnera}
\author[inst2]{Pasquale Caponnetto}
\author[inst1]{Sebastiano Battiato}

\affiliation[inst1]{organization={Department of Mathematics and Computer Science, University of Catania},
            addressline={Viale Andrea Doria 6}, 
            city={Catania},
            postcode={95126}, 
            state={CT},
            country={Italy}}

\affiliation[inst2]{organization={Department of Educational Sciences, Section of Psychology, University of Catania},
            addressline={Via Teatro Greco 84}, 
            city={Catania},
            postcode={95124}, 
            state={CT},
            country={Italy}}

\cortext[cor1]{mirko.casu@phd.unict.it; Department of Mathematics and Computer Science, University of Catania, Viale Andrea Doria 6, Catania, 95126, CT, Italy}

\begin{abstract}
This paper examines the impact of cognitive biases on decision-making in forensics and digital forensics, exploring biases such as confirmation bias, anchoring bias, and hindsight bias. It assesses existing methods to mitigate biases and improve decision-making, introducing the novel “Impostor Bias", which arises as a systematic tendency to question the authenticity of multimedia content, such as audio, images, and videos, often assuming they are generated by AI tools. This bias goes beyond evaluators' knowledge levels, as it can lead to erroneous judgments and false accusations, undermining the reliability and credibility of forensic evidence. Impostor Bias stems from an a priori assumption rather than an objective content assessment, and its impact is expected to grow with the increasing realism of AI-generated multimedia products. The paper discusses the potential causes and consequences of Impostor Bias, suggesting strategies for prevention and counteraction. By addressing these topics, this paper aims to provide valuable insights, enhance the objectivity and validity of forensic investigations, and offer recommendations for future research and practical applications to ensure the integrity and reliability of forensic practices.
\end{abstract}




\end{frontmatter}


\section{Introduction}
\label{sec:introduction}


In forensic sciences, the objectivity of judgment in analyzing data for justice purposes is paramount. Beyond technical expertise, awareness of cognitive biases is crucial.  These biases, rather than being deficits, are systematic preferences that influence the way we process, select, and retain information~\citep{lester2011modifying,grisham2014using}. They can have both positive and negative effects, depending on the context and situation, facilitating swift decision-making when time is critical but also leading to poor decisions and adverse outcomes~\citep{meterko2022cognitive,berthet2022impact}.

Some of these biases have been identified as a significant factor impacting the objectivity and accuracy of forensic science \citep{bhadra2021forensic}. For example, law enforcement professionals showed vulnerable to \textit{confirmation bias} \citep{meterko2022cognitive}, which is a tendency to search for, interpret, and remember information in a way that confirms one’s preexisting beliefs or hypotheses \citep{nickerson1998confirmation}. Another recurring bias in forensics is the \textit{anchoring bias} or \textit{effect} \citep{Edmond_Tangen_Searston_Dror_2015}, which occurs when an individual relies too heavily on an initial piece of information (the “anchor”) when making decisions \citep{Chapman1994The,Chapman1999Anchoring}. One more relevant bias related to forensic science is the \textit{hindsight bias} \citep{Giroux2016Hindsight}, which occurs when people believe that an event is more predictable after it becomes known, involving memory distortion, beliefs about objective likelihoods, and subjective beliefs about one's own prediction abilities \citep{Roese2012Hindsight}. The discipline of digital forensics is not exempt from these challenges \citep{sunde2019cognitive}. Consequently, an increasing number of scholarly investigations are focusing on this particular field \citep{Sunde_Dror_2021}.

This paper explores the impact of cognitive biases in forensics and digital forensics, with a focus on deepfakes and Artificial Intelligence (AI)-generated multimedia content, which pose threats such as manipulating public opinion and impersonating individuals. We introduce the \textit{Impostor Bias}, an inherent distrust of multimedia authenticity due to the prevalence of AI-generated content. Effective detection of deepfakes is crucial to prevent confusion between real and fake multimedia, which can lead to erroneous decisions. We analyze advanced deepfake detection systems to aid operators in distinguishing authentic content, addressing the challenges posed by deepfakes, and ensuring accurate multimedia evaluation in digital forensics.

Particularly, the following points encapsulate the salient findings of this article:

\begin{itemize}
\item Cognitive biases in digital forensics: we discuss how cognitive biases can affect the perception and judgment of digital forensic investigators, especially in the face of complex and large-scale data.
\item Deepfake detection methods: the state-of-the-art methods for detecting deepfakes, which are synthetic media created by advanced AI technologies, such as GANs and DMs, are reviewed.
\item The Impostor Bias: we unveil the new concept of the Impostor Bias, which is the tendency to doubt the authenticity of real media due to the proliferation of deepfakes and the difficulty of distinguishing them from reality.
\item Biases mitigating strategies: some strategies to reduce the impact of cognitive biases in digital forensics, such as using objective and standardized procedures, are proposed to enhance the training and education of forensic experts, and to adopt ethical and legal guidelines.
\end{itemize}

Finally, the paper is structured as follows: Section 1 introduced the concept of cognitive biases. Section 2 analyses their impact on forensic sciences. Section 3 explores some examples of cognitive biases in digital forensics, such as confirmation bias and pareidolia bias. Section 4 explores various bias mitigation strategies in both forensics and digital forensics. Section 5 presents the deepfakes and how they can be generated and managed. Section 6 introduces the Impostor Bias, a new type of bias triggered by AI media that affects the perception of reality. Section 7 reviews some of the most recent and relevant methods for deepfake detection, which is crucial to counter the Impostor Bias, as well as the problem of model attribution. Section 8 discusses the potential impact of Impostor Bias in digital forensics and everyday life. Section 9 concludes the paper and provides some directions for future research.

\section{Cognitive Biases Impact on Forensic Sciences}
\label{sec:biasfor}

Cognitive biases have been a concern in forensic science since 1984, when Larry Miller published a work discussing the presence of bias in forensic document examiners~\citep{Stoel2014Bias}. He suggested the introduction of procedural modifications to reduce cognitive bias, which could potentially result in incorrect outcomes.

These biases can significantly impact expert judgments and the criminal justice process~\citep{Stoel2014Bias, dror2008meta, Neal2014The}, are influenced by various factors, and can lead to errors and misinterpretation of evidence~\citep{kassin2013forensic, cooper2019cognitive, bhadra2021forensic}. Both forensic experts and law enforcement professionals are susceptible to these biases, and one of the most common of them is the \textit{confirmation bias}~\citep{Moser_2013,Thompson2015Lay,van2019forensic,cooper2019cognitive,meterko2022cognitive}: if a forensic analyst believes a suspect is guilty, they might unconsciously interpret ambiguous evidence as incriminating \citep{kassin2013forensic,van2019forensic}.~\citet{gardner2019evidence} found that task-irrelevant information can bias forensic analysts' decisions, suggesting that extraneous details can inadvertently sway the interpretation of evidence. This is echoed by ~\citet{nakhaeizadeh2014cognitive}, who discovered that irrelevant information can lead to confirmation bias in forensic anthropology, potentially leading to skewed conclusions based on pre-existing beliefs rather than objective evidence.~\citet{dror2021cognitive} further explored this issue, finding that base-rate neglect could bias forensic pathologists' decisions in child death cases. This suggests that statistical information about the prevalence of certain causes of death may be overlooked, leading to potential misinterpretations. In the realm of DNA forensics,~\citet{Jeanguenat2017Strengthening} found that suspect-driven bias and the presence of rare alleles can influence interpretation. This highlights the potential for preconceived notions about a suspect, as well as the rarity of certain genetic markers, to affect the analysis of DNA evidence. In a study by \citet{douglass2023case}, evaluators successfully differentiated accurate from inaccurate witnesses based on videos of identification procedures alone, but their ability to discern accuracy was disrupted when extraneous incriminating evidence was also provided. This aligns with confirmation bias, where evaluators tend to favor information that confirms their existing beliefs or expectations, even when it contradicts objective evidence. Regarding the \textit{anchoring bias}, in a forensic context, an analyst might give undue weight to the first piece of evidence they examine, which could skew their interpretation of subsequent evidence \citep{Edmond_Tangen_Searston_Dror_2015,meterko2022cognitive}. For the \textit{hindsight bias}, this could lead to overconfidence in the accuracy of a forensic analysis after a suspect has been identified \citep{Giroux2016Hindsight,Beltrani2018Is,meterko2022cognitive}.~\citet{Neal2022A} conducted a systematic review on cognitive biases and debiasing techniques in forensic mental health, finding significant bias effects specifically for confirmation and hindsight bias. Lastly,~\citet{stevenage2017biased} found that irrelevant DNA test outcomes could bias fingerprint matching tasks, confirming the presence of contextual bias. This suggests that unrelated test results can influence the interpretation of fingerprint evidence.

\begin{figure}[t!]
    \centering
    \includegraphics[width=1\linewidth]{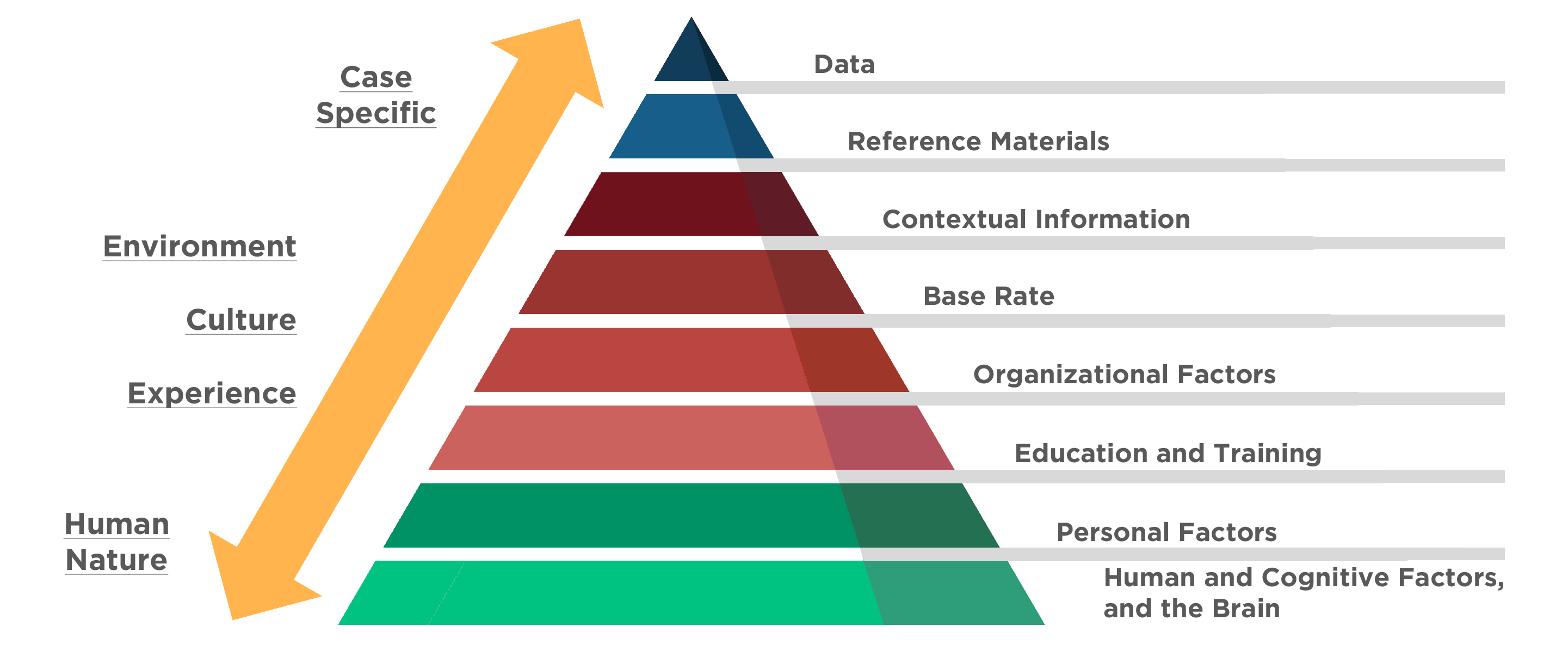}
    \caption{Eight potential sources of bias that could impact forensic decision-making.}
    \label{fig:pyramidbias}
\end{figure}

The sources of bias can be categorized into three groups related to the case, the analyst, and human nature. Inspired by the work of Dror et al.~\citep{dror2020cognitive}, we define a taxonomy of potential sources of bias. These biases can introduce cognitive distortions into the processes of sampling, observing, strategizing tests, analyzing, and drawing conclusions, even when conducted by experts. This taxonomy is synthetically sketched in Figure~\ref{fig:pyramidbias}.

\section{Exploring Cognitive Biases in Digital Forensics}
\label{sec:biasdigfor}

The definition of Digital Forensic Science often referred to is the one from the Digital Forensic Research Workshop (DFRWS) in~\citet{palmer2001road}: “The use of scientifically derived and proven methods toward the preservation, collection, validation, identification, analysis, interpretation, documentation and presentation of digital evidence derived from digital sources for the purpose of facilitating or furthering the reconstruction of events found to be criminal, or helping to anticipate unauthorized actions shown to be disruptive to planned operations”. A key area within this field involves the proper acquisition of digital content such as images, videos, and audio to produce evidence for forensic investigations. Multimedia forensics focuses on verifying the authenticity of data and reconstructing the history of an image since its acquisition~\citep{battiato2016multimedia,fabio2023innovative,giudice2017classification,piva2013overview}.

In the realm of digital forensics, cognitive bias emerges as a subtle yet potent force that can shape the outcomes of investigations. \citet{Sunde_Dror_2021} explored the susceptibility of digital forensics examiners to bias, revealing how preconceived notions and contextual information can skew their observations and interpretations. Despite the digital evidence's facade of objectivity, the human factor introduces variability, leading to inconsistent conclusions among experts analyzing identical datasets \citep{Sunde_Dror_2021}.

\subsection{Confirmation Bias in Text and Face Recognition}
\label{sub:textfacerec}

\begin{figure}[t!]
    \centering
    \includegraphics[width=1\linewidth]{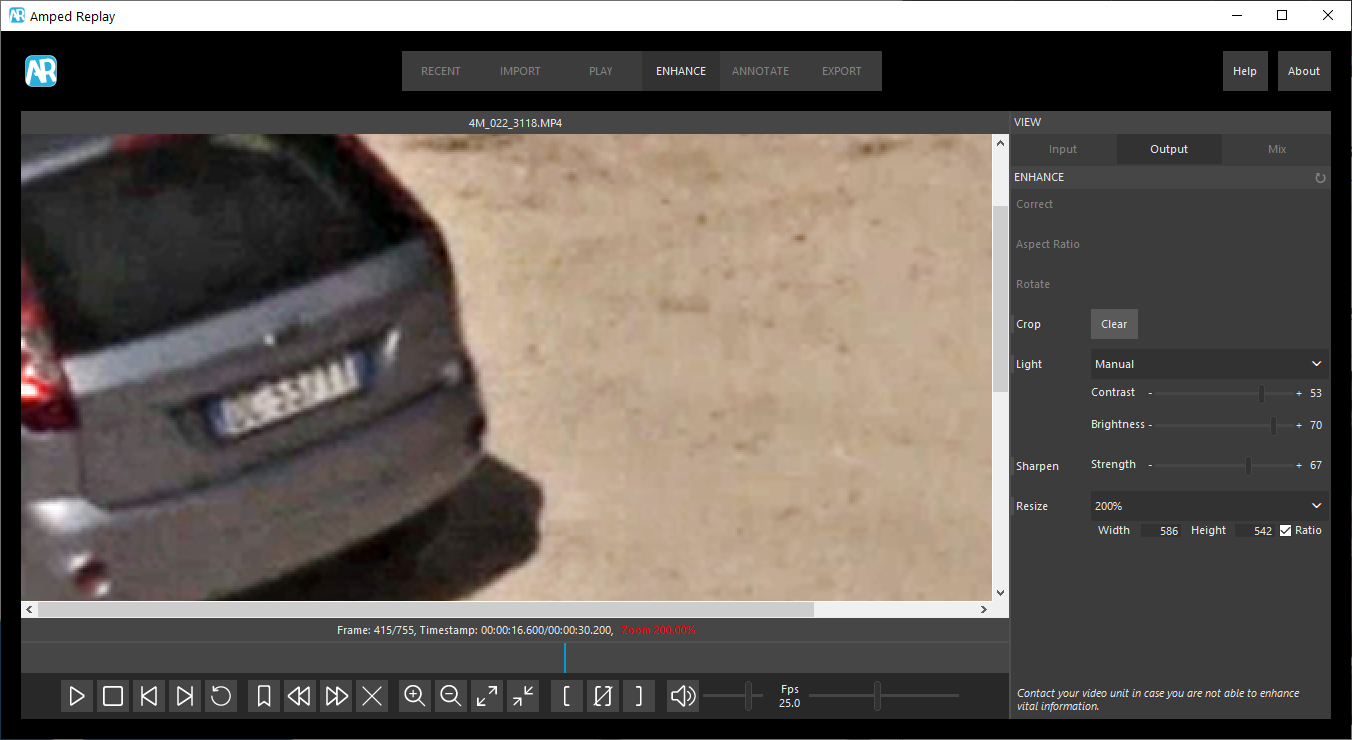}
    \caption{John asked Lucy for help with a license plate, reading “BC 537”, but unsure about the last two characters and the first one. Credit:~\citet{fontani_cognitive_2021}.}
    \label{fig:licenseplate}
\end{figure}

\begin{figure}[t!]
    \centering
    \includegraphics[width=1\linewidth]{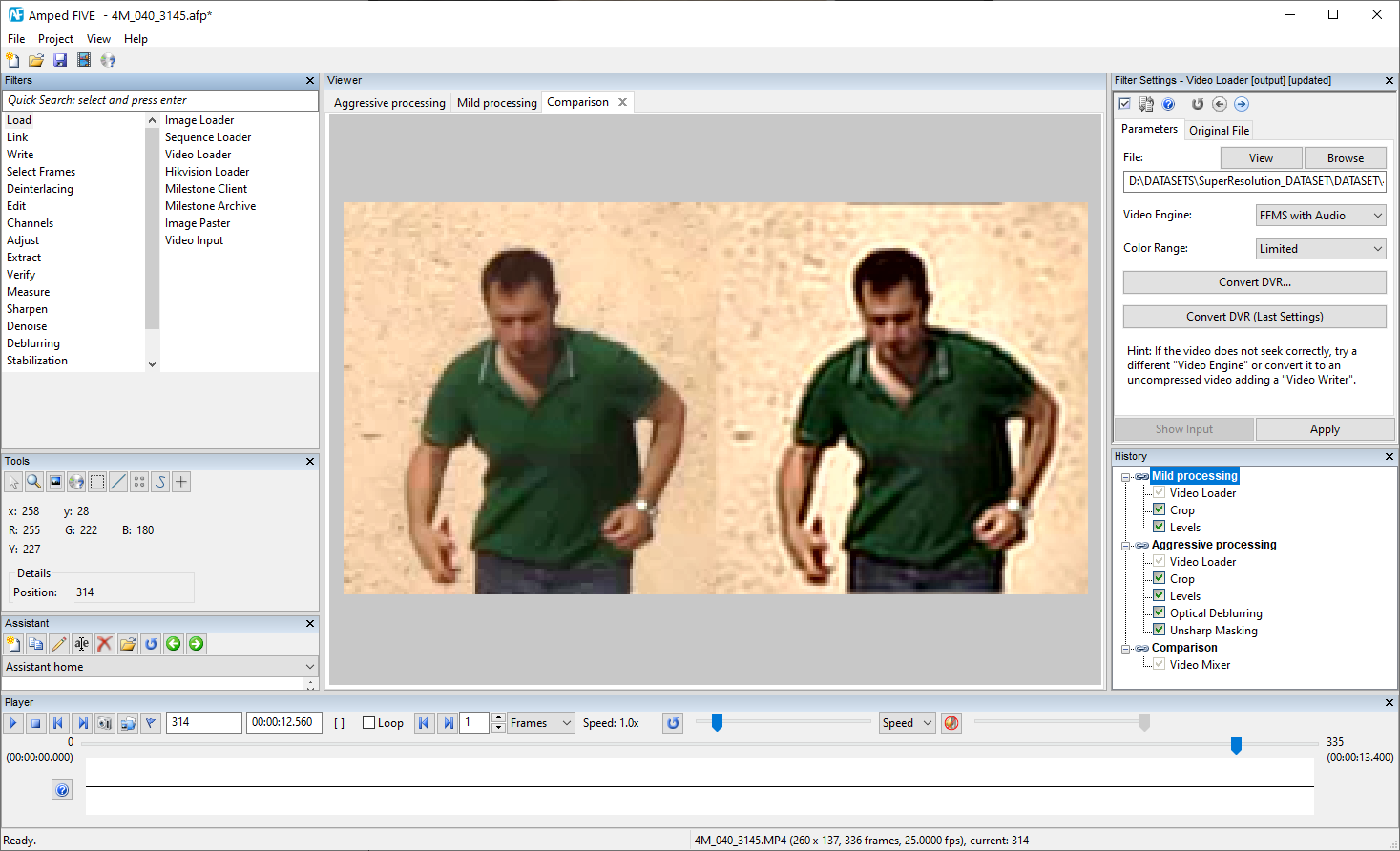}
    \caption{Image processing could lead to produce a noticeable different image, with different face characteristics. Credit:~\citet{fontani_cognitive_2021}.}
    \label{fig:sharperimage}
\end{figure}

Fontani's blog post on Amped Software depicted a hypothetical situation featuring two characters, John and Lucy (Figure~\ref{fig:licenseplate})~\citep{fontani_cognitive_2021}. In this scenario, John inadvertently influenced Lucy's interpretation of a license plate by prematurely sharing his own interpretation. The cognitive bias present in this example could be the \textit{confirmation bias}, since this latter is a type of cognitive bias where individuals are more likely to seek out, interpret, and remember information that confirms their pre-existing belief \citep{nickerson1998confirmation,cooper2019cognitive,meterko2022cognitive}. John's premature sharing of his interpretation could lead Lucy to interpret the license plate in the same way, confirming John's interpretation rather than considering other possible interpretations. It may lead Lucy to unconsciously process pixels and select frames that align with John's interpretation. The post further exploited into the intricacies of face comparison, noting that varying processing techniques can result in significantly different facial appearances. It underscored the necessity of withholding the suspect’s face from the examiner prior to the enhancement process to prevent unconscious bias towards a match. Fontani also emphasized that different processing techniques can significantly alter facial appearances during enhancement (Figure~\ref{fig:sharperimage}). Therefore, it's crucial that the examiner doesn't see the suspect’s face before the enhancement process to avoid unconsciously adjusting the enhancement to create a match.

Furthermore,~\citet{sunde2019cognitive} underscored the importance of digital forensics as a rapidly growing field within forensic science. The authors analyzed seven specific sources of cognitive and human error within the digital forensics process and propose relevant countermeasures, concluding that while some cognitive and bias issues are common across forensic domains, others are unique and dependent on the specific characteristics of the domain, such as digital forensics.

\subsection{Image Processing Could Lead to Pareidolia}
\label{sub:pareidolia}

\begin{figure}[t!]
    \centering
    \begin{subfigure}{.4\textwidth}
        \centering
        \includegraphics[width=0.7\linewidth]{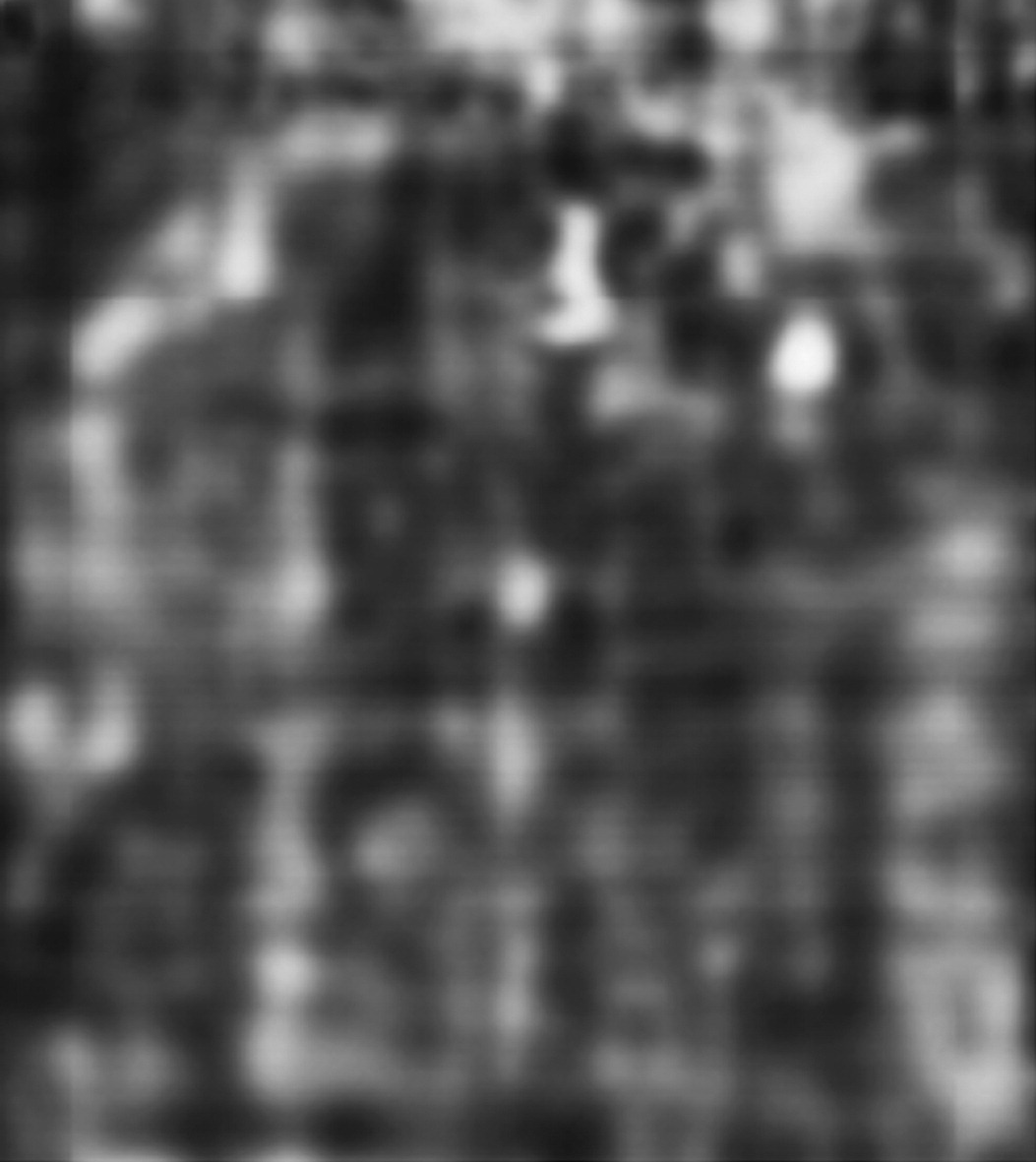}
        \caption*{(a)}
    \end{subfigure}%
    \begin{subfigure}{.4\textwidth}
        \centering
        \includegraphics[width=0.7\linewidth]{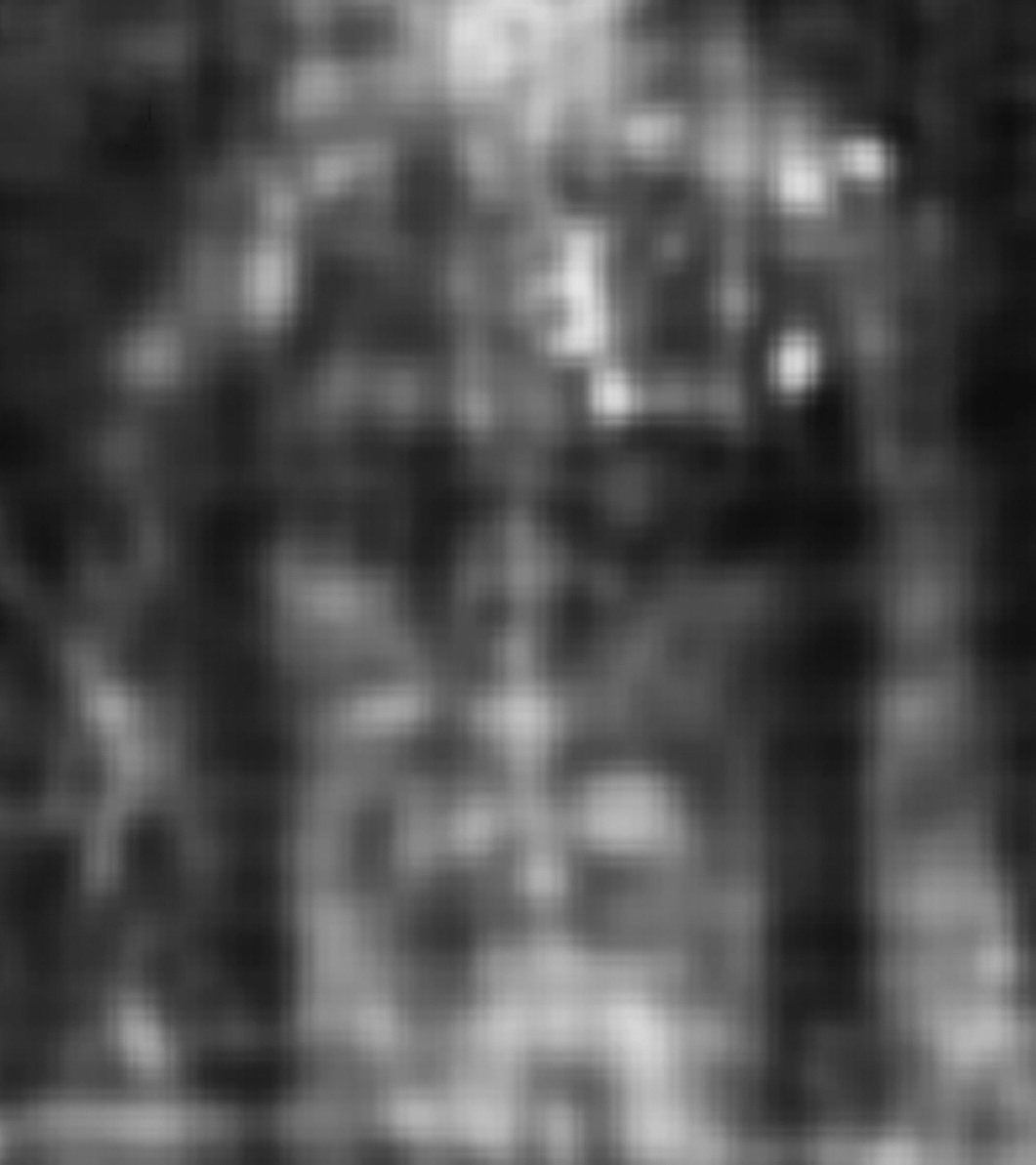}
        \caption*{(b)}
    \end{subfigure}
    \caption{A supposed concealed image of a face on the back side of the Shroud is revealed through advanced image processing of a photograph published in a book. The image is flipped from right to left (b). A negative image of the face that can be seen on the front side of the Shroud, processed in the same way as (a). Credit:~\citet{fanti2004double}.}
    \label{fig:shroud1}
\end{figure}

\begin{figure}[t!]
    \centering
    \begin{subfigure}{.4\textwidth}
        \centering
        \includegraphics[width=0.6498\linewidth]{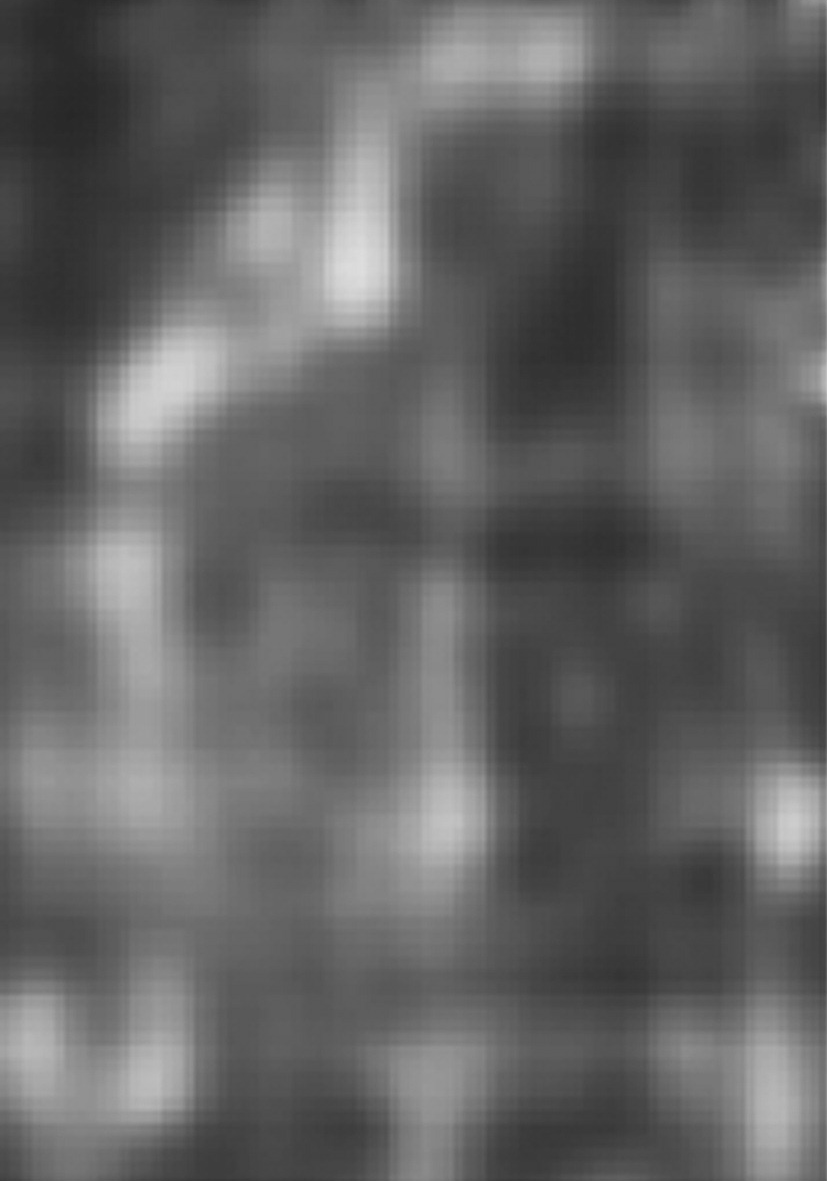}
        \caption*{(a)}
    \end{subfigure}%
    \begin{subfigure}{.4\textwidth}
        \centering
        \includegraphics[width=0.7553925\linewidth]{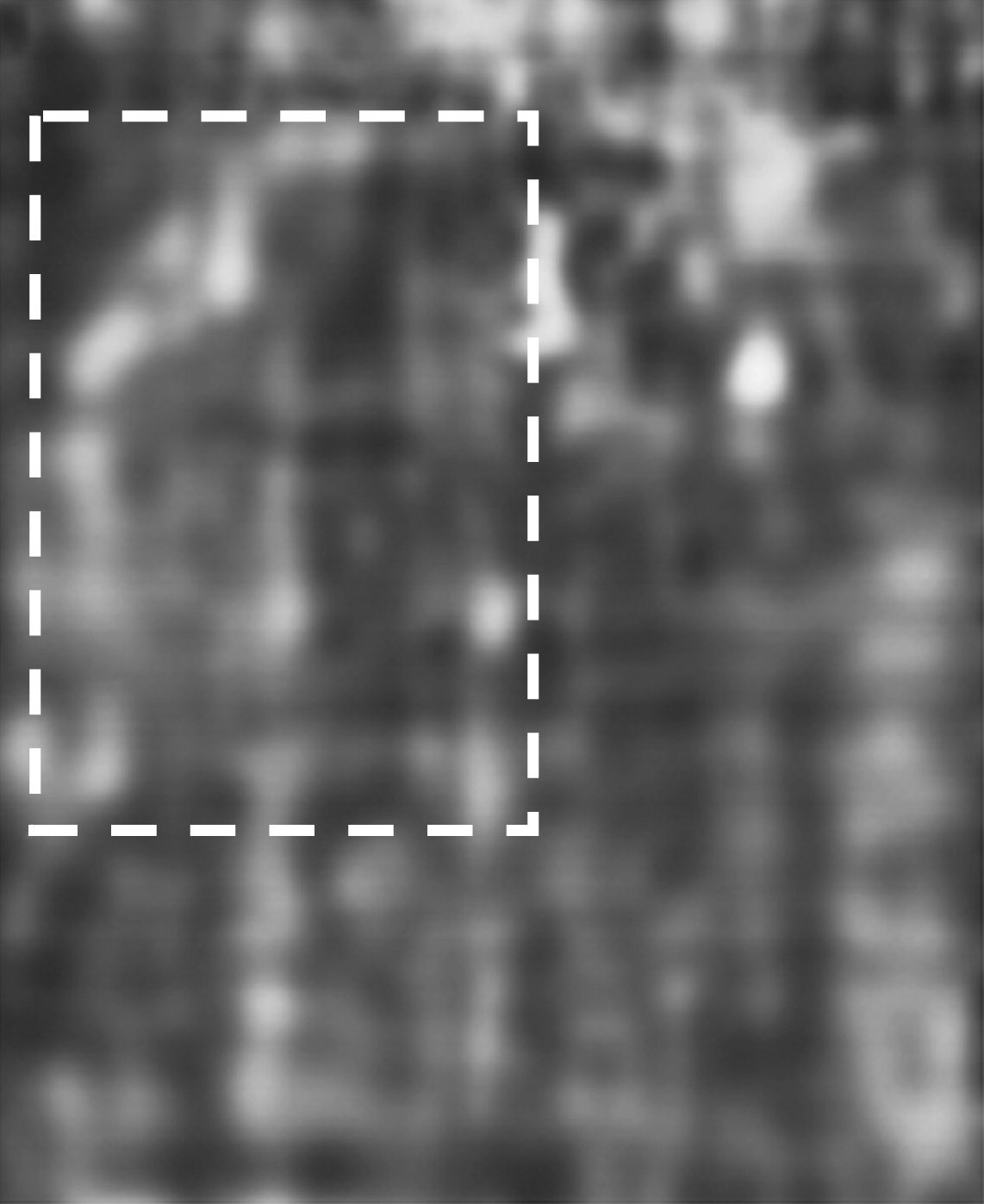}
        \caption*{(b)}
    \end{subfigure}
    \caption{A magnified version of Figure~\ref{fig:shroud1}b (a). A face resembling the Shroud that we discern in the top-left section of Figure~\ref{fig:shroud1}b (as depicted on the right) (b). We can also discern another face in the bottom left section of Figure~\ref{fig:shroud1}b. Pareidolia leads to false positives, enabling us to see faces in Figure~\ref{fig:shroud1}b that aren't actually there. Credit:~\citet{di2013pattern}.}
    \label{fig:shroud2}
\end{figure}

A study by~\citet{di2013pattern} focused on the potentially misleading effects of software techniques used for elaborating low-contrast images. The researchers used the Shroud of Turin, one of the most studied archeological objects in history, as an example (Figure~\ref{fig:shroud1}). They demonstrated that image processing of both old and recent photographs of the Shroud could lead researchers to perceive inscriptions and patterns that do not actually exist. The study further emphasized that the limited static contrast of our eyes can make the perception of low-contrast images problematic. The brain's ability to retrieve incomplete information can interpret false image pixels after image processing. This phenomenon, named “pareidolia”, can lead to the perception of patterns in Shroud photographs that do not exist in reality (Figure~\ref{fig:shroud2}).

The enhancement of images extracted from video cameras can indeed lead to a degradation of the overall quality of information. This degradation can be attributed to factors such as excessive compression \citep{maity2023survey}, distance from the recording plane \citep{wang2023cham}, and limited overall resolution \citep{maity2023survey,wang2023cham}. International best practices suggest verifying on a case-by-case basis whether the level of information is sufficient to extract useful data for investigations \citep{tenopir2020,Soltani_Nikou_2020}.

However, when examining low-contrast images that present pseudo-random visual patterns after an initial enhancement process, it is crucial to mitigate the risk of pareidolia. This bias is particularly potent when the object of interest refers to “human faces” or more generally to “letters/numbers” or known human structures~\citep{Wang2018Face,Zhou2020Do}. Pareidolia is a subconscious illusion that tends to associate random shapes with known forms, especially human figures and faces. Classic examples include seeing animals or human faces in clouds, or a human face on the moon. In forensic investigations, we also suggest to entrust the analysis and interpretation to automatic methods \citep{Solanke_Biasiotti_2022} or experts who can follow a “blind testing” approach, i.e., an interpretation detached from the knowledge of details and the reference context \citep{Cowan_Koppl_2011,Servick_2015}. This helps to ensure the search for “certain” evidence is as rigorous and unbiased as possible.

~\citet{Zhou2020Do} discussed how some individuals exhibit a looser decision criterion for detecting faces, making them more prone to perceive faces where none exist. This relates to a concept in signal detection theory known as response bias~\citep{Nguyen2013Response}. In digital forensics, examiners may fall prey to response biases when analyzing ambiguous digital evidence, predisposing them to validate or dismiss forensic hypotheses based on non-diagnostic features. Just as some are more likely to see faces in random patterns due to biases in how they set thresholds for face judgments, forensic analysts could have biases influencing how strictly they apply standards of evidence to digital artifacts. Understanding individual differences in cognitive biases like threshold placement could help address potential sources of error and increase objectivity in digital forensic examinations \citep{Berthet_2021,Horsman_2024}.

\subsection{Case Study: Confirmation Bias and Pareidolia in Surveillance Camera Footage}
\label{sub:casestudy}

\begin{figure}[b!]
    \centering
    \includegraphics[width=1\linewidth]{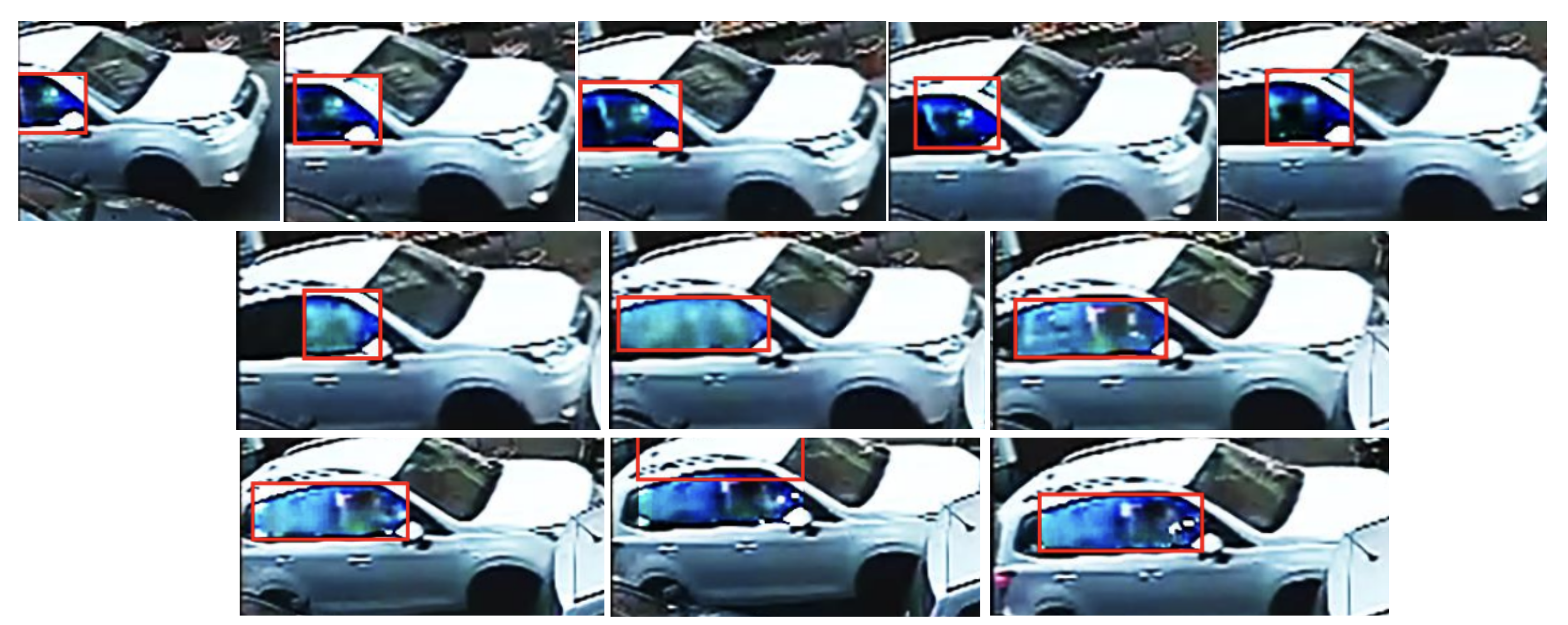}
    \caption{Sequence of frames that have been processed by the technical consultant.}
    \label{fig:finestrinoLebano}
\end{figure}

We present a case study with the objective of determining the presence of a passenger in a vehicle involved in a murder criminal case, through the analysis of surveillance camera footage. The methodology included the scrutiny of various cameras and a detailed analysis of the vehicle's passages from different surveillance cameras, considering image overlaps and real passage times, including sunset. The data for this study is primarily derived from the surveillance camera footage. Figure \ref{fig:finestrinoLebano} shows the only sequence of frames (among all the collected surveillance footage) in which the passenger's presence is doubtful, and that has been processed by a technical consultant. These frames demonstrate a clear reflection effect that disappears as the vehicle moves, and exhibit strong chromatic variability due to video compression. The technical advisor claimed to discern a human face in these images, a perception \textit{potentially} induced by confirmation bias, as his intervention was required precisely to identify the possible presence of the passenger. Specifically, on the passage of the car identified by the technical advisor he identifies with certainty, in the first frames of the passage, a face as a “clear” silhouette traceable to a passenger placed in the right front seat. A clear spot can be seen that he traces back to a silhouette. This spot disappears completely in the second part of the passage and then reappears but in inverted colors, that is, it becomes dark.  In reality, the images contain pseudo-random blobs, and their temporal persistence can be attributed to a simpler and more evident reflection. The study highlights the lack of scientific rigor in these approaches and proposes new analyses that suggest the absence of a passenger in the vehicle.

\begin{table}[t!]
\centering
\resizebox{0.8\textwidth}{!}{%
\begin{tabular}{cl}
\hline
\toprule
\textbf{Image} &
  \multicolumn{1}{c}{\textbf{Students' Evaluation}} \\ \midrule                                                                                  
\raisebox{-0.5\height}{\includegraphics[width=0.30\textwidth]{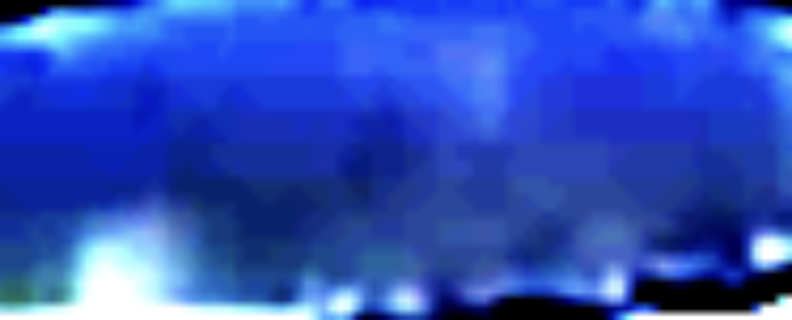}} & \begin{tabular}[c]{@{}l@{}}Nothing: 8 Other: 7\\ Those who answered “other" identified:\\ • Automobile elements\end{tabular}                                              \\ \hline
\raisebox{-0.5\height}{\includegraphics[width=0.30\textwidth]{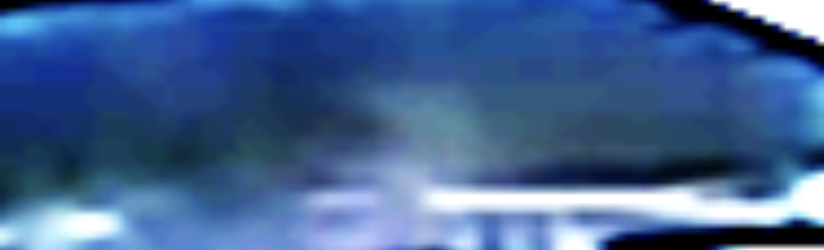}} & \begin{tabular}[c]{@{}l@{}}Nothing: 9 Other: 6\\ Those who answered “other" identified:\\ • Automobile elements\end{tabular}                                              \\ \hline
\raisebox{-0.5\height}{\includegraphics[width=0.30\textwidth]{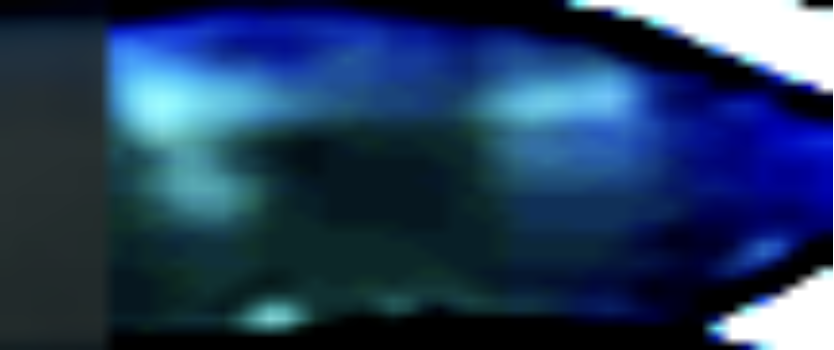}} & \begin{tabular}[c]{@{}l@{}}Nothing: 10 Other: 5\\ Those who answered “other" identified:\\ • Silhouette with sun reflection\\ • Automobile elements\end{tabular}          \\ \hline
\raisebox{-0.5\height}{\includegraphics[width=0.30\textwidth]{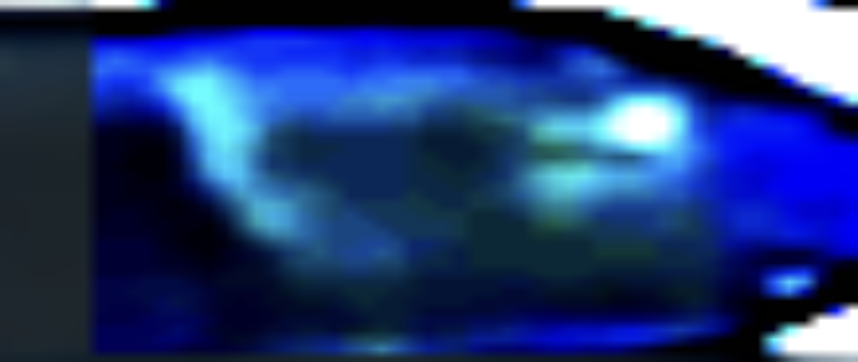}} & \begin{tabular}[c]{@{}l@{}}Nothing: 9 Other: 6\\ Those who answered “other" identified:\\ • A car seat and a person\\ • Reflected human silhouette\\ • A cow\end{tabular} \\ \hline
\raisebox{-0.5\height}{\includegraphics[width=0.30\textwidth]{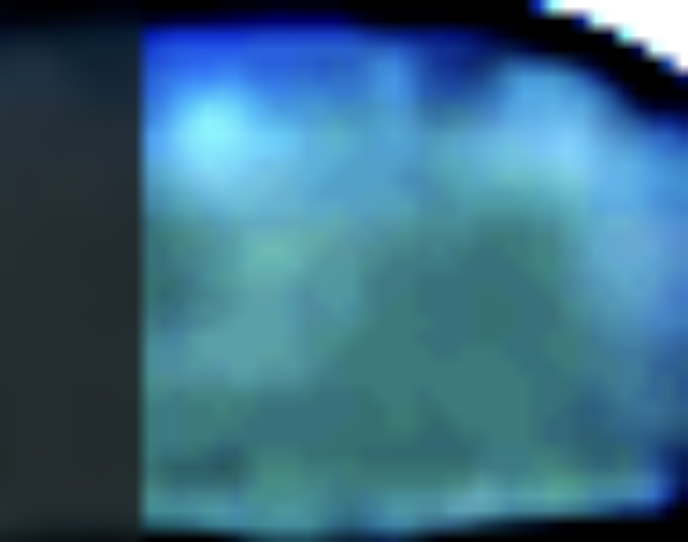}} & \begin{tabular}[c]{@{}l@{}}Nothing: 13 Other: 2\\ Those who answered “other" identified:\\ • Indistinct silhouettes behind glass\end{tabular}                             \\ \hline
\raisebox{-0.5\height}{\includegraphics[width=0.30\textwidth]{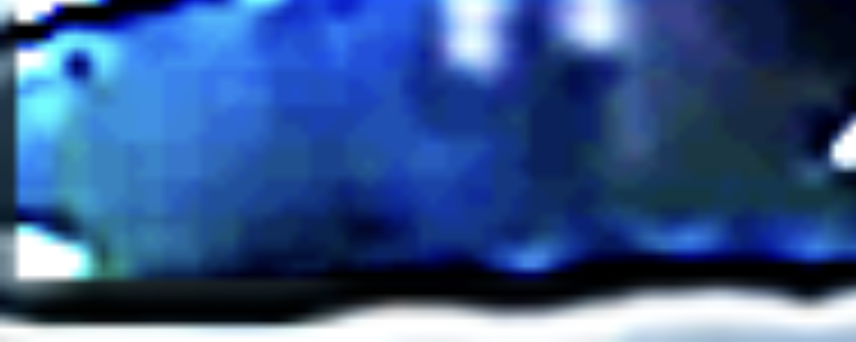}} & \begin{tabular}[c]{@{}l@{}}Nothing: 9 Other: 6\\ Those who answered “other" identified:\\ • There are two human silhouettes\\ • Silhouettes of hands\end{tabular}         \\ \hline
\end{tabular}
}
\caption{Table representing students' evaluation of the various images extracted from Figure~\ref{fig:finestrinoLebano}.}
\label{tab:students}
\end{table}

An experiment was conducted involving college students. The students were anonymously and without context presented with carefully selected frames from the surveillance camera footage, focusing on the most controversial and doubtful portions. The images shown to the students were a representative subset of the entire set of images extracted and processed by the technical advisor (Figure\ref{fig:finestrinoLebano}). To ensure impartiality in their observations, no preliminary information was provided to the students. As part of the experiment, a questionnaire was administered for each image, consisting of the question: “What do you observe in this image?", with response options such as “Nothing" and “Other," along with an open text field for further clarification. This design encouraged a spectrum of responses and allowed students to give concise yet descriptive explanations. A total of 15 subjects participated in evaluating the selected images, resulting in 165 responses, as summarized in Table~\ref{tab:students}. Notably, only two of the responses conveyed a "certain" identification of a human face. This underscores the necessity of meticulous and unbiased analysis when interpreting surveillance footage. The primary objective of the experiment was to assess the students' ability to detect the presence of a passenger in the vehicle. By presenting them with ambiguous visuals, the experiment aimed to understand their perceptual limitations and biases. The absence of explicit certainty measurement in the questionnaire was addressed by considering the students' chosen responses as an indication of their certainty levels. This highlights the importance of rigorous and unbiased analysis in interpreting surveillance footage. The study emphasizes the need to maintain an objective perspective in observational tasks \citep{Altmann1974Observational} and underscores the value of scientific rigor in such analyses.


\section{Strategies for Mitigating Cognitive Bias}
\label{sec:mitstr}

Among other fields, software engineering is currently experiencing a significant gap in the area of cognitive bias mitigation techniques, with a notable lack of both practical strategies and theoretical foundations \citep{Mohanani2017Cognitive}. However, other fields have seen success in this area. For instance, in social work, the use of a nomogram tool and an online training course has been shown to effectively mitigate cognitive bias, leading to improvements in the accuracy of clinical reasoning \citep{Featherston_Shlonsky_Lewis_Luong_Downie_Vogel_Granger_Hamilton_Galvin_2019}. Despite these advancements, it's important to note that there is currently insufficient evidence to suggest that cognitive bias mitigation interventions significantly improve decision-making in real-life situations \citep{korteling2021retention}.

Nevertheless, the potential benefits of countering cognitive biases are clear. In healthcare, for example, addressing cognitive biases in decision-making can help reduce low-value care and enhance the impact of campaigns aimed at reducing such care \citep{Scott_Soon_Elshaug_Lindner_2017}. In the realm of gaming, the MACBETH serious game has been found to effectively mitigate cognitive biases such as the fundamental attribution error \citep{Miller1989The} and confirmation bias. The game's effectiveness is further enhanced through explicit instruction and repetitive play, which serve to reinforce learning \citep{Dunbar_Miller_Adame_Elizondo_Wilson_Lane_Kauffman_Bessarabova_Jensen_Straub_etal._2014}.

\subsection{Strategies for Mitigating Cognitive Bias in Forensic Science}
\label{sub:mitstrfor}

In the field of forensic science, cognitive biases can significantly impact the accuracy and impartiality of examiner decisions. Specifically,~\citet{Dror2013Practical} focused on strategies to mitigate confirmation bias, contextual influences \citep{nakhaeizadeh2014cognitive}, and base-rate regularities \citep{Thakur_Basso_Ditterich_Knowlton_2021}. Dror proposed that recognizing the spectrum of biases, not only those that can arise from knowing irrelevant case information, but also biases that emerge from base rate regularities, working “backwards” from the suspect to the evidence, and from the working environment itself, can strengthen forensic science.

To mitigate these effects, several strategies have been proposed. First, cognitive training programs raise awareness among examiners about potential biases. Second, blind verification procedures, where the second examiner is unaware of the first examiner's decision, help minimize bias. Third, linear examination processes, starting with evidence analysis before considering the suspect, reduce contextual contamination. Fourth, a triage approach tailors procedures based on case complexity, ensuring that resources are allocated effectively. Lastly, cognitive profiles aid in selecting the best-suited individuals for forensic work, enhancing overall objectivity and performance.

In the pursuit of objectivity and accuracy in forensic analysis, the Forensic Science Regulator of the United Kingdom Government has proposed and implemented several strategies designed to ensure that the analysis is not influenced by any form of bias~\citep{ForensicScienceRegulator2020}.

\begin{enumerate}
    \item Blinding Precautions:
    \begin{enumerate}
        \item Analysts should be shielded from information that is not directly relevant to the analysis.
        \item Sequential unmasking can be used, where decisions on suitability are made before comparison with reference samples.
        \item Careful records should be kept to ensure the order of disclosure and analysis is transparent.
    \end{enumerate}
    \item Structured Approach (ACE-V and CAI):
    \begin{enumerate}
        \item ACE-V (Analysis, Comparison, Evaluation, and Verification) provides a structured process for fingerprint comparison~\citep{reznicek2010ace}.
        \item CAI (Case Assessment and Interpretation) uses Bayesian thinking and balances prosecution and defense hypotheses~\citep{jackson2011development}.
        \item Both approaches emphasize transparency and avoid post hoc rationalization.
    \end{enumerate}
    \item Awareness, Training, and Competence Assessment:
    \begin{enumerate}
        \item Practitioners need training on cognitive bias risks and mitigation strategies.
        \item Proficiency testing and regular assessment help maintain competence.
    \end{enumerate}
    \item Avoidance of Reconstructive Effects:
    \begin{enumerate}
        \item Contemporaneous notes or technical records prevent reconstructive bias.
        \item Analysts should rely on memory as little as possible.
    \end{enumerate}
    \item Avoidance of Role Effects:
    \begin{enumerate}
        \item Organizational structures should insulate scientists from potential biasing pressures.
        \item Scientists must prioritize their duty to the court over any other obligations.
    \end{enumerate}
\end{enumerate}

Controlling the flow of information to analysts is crucial to prevent unnecessary influences on their judgment. \citet{dror2021linear} introduced Linear Sequential Unmasking–Expanded (LSU-E), a methodology that reduces noise and bias in forensic decision-making. LSU-E involves initial analysis of raw data without reference material, followed by a sequential consideration of relevant information based on objectivity and relevance. This approach optimizes information presentation to enhance utility and minimize cognitive biases. LSU-E also offers guidelines for documenting the influence of information on the decision-making process, ensuring transparency and accountability. Furthermore, \citet{Camilleri2019A} emphasized the need for a systematic assessment of cognitive bias risks in forensic laboratories, proposing a risk management framework. Key mitigation strategies include raising awareness through training, developing guidance documents, and limiting access to task-irrelevant information. Redacting irrelevant details, implementing blind known tests, and conducting independent casefile reviews enhance objectivity and reduce expectation bias. These measures may ensure the integrity of forensic interpretations and minimize the impact of cognitive biases.

\section{The Intersection of Generative AI and the Craftsmanship of Deepfakes}
\label{sec:deepfakes}

Generative Artificial Intelligence (GenAI) is an increasingly popular technology that has significant implications across various fields \citep{Bockting2023,Stokel-Walker_VanNoorden_2023}. It refers to AI systems that can generate new content, such as text, images, and audio, in response to human prompts. These systems, including deepfakes and AI chatbots like Generative Pre-trained Transformers (e.g., GPT-4), use complex algorithms to produce outputs that are often indistinguishable from content created by humans \citep{Stokel-Walker_VanNoorden_2023}. The technology is advancing rapidly, with each new version adding capabilities that increasingly encroach on human skills; however, the use of these “black box" AI tools can introduce biases and inaccuracies, potentially distorting scientific facts while still sounding authoritative.

\begin{figure}[b!]
    \centering
    \includegraphics[width=1\linewidth]{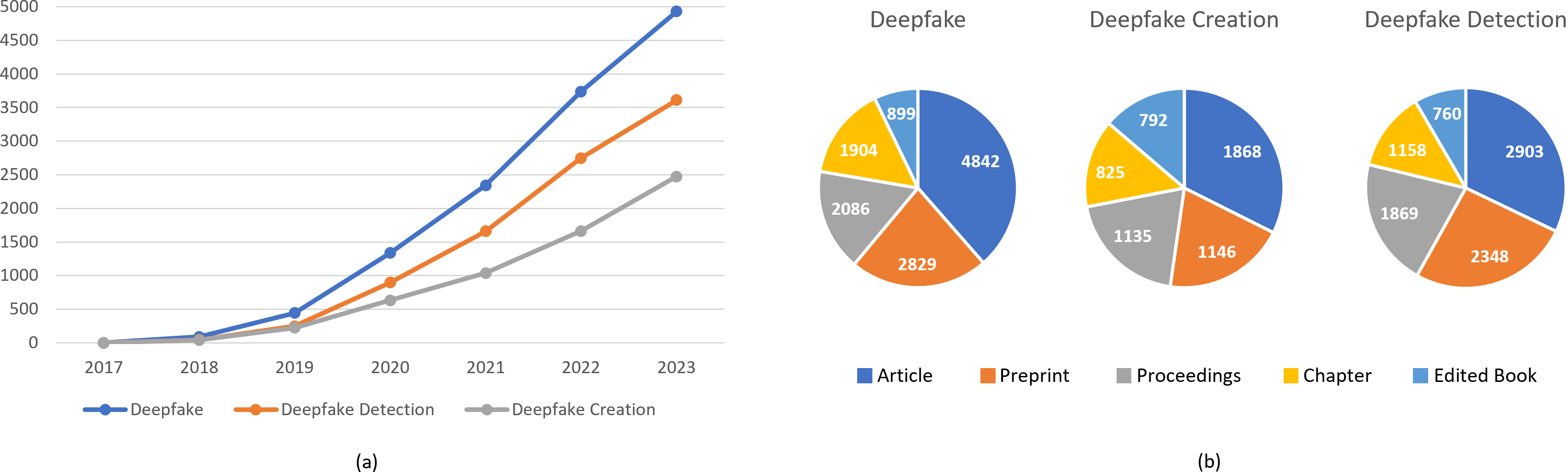}
    \caption{Statistics of papers published in the deepfake field. (a) Papers published from 2017 to 2023 with the keywords deepfake, deepfake creation, deepfake detection. (b) Numbers of papers published in: Article, Preprint, Proceedings, Chapter and Edited Book.}
    \label{fig:paperstatistics}
\end{figure}

\begin{figure}[b!]
    \centering
    \includegraphics[width=.8\linewidth]{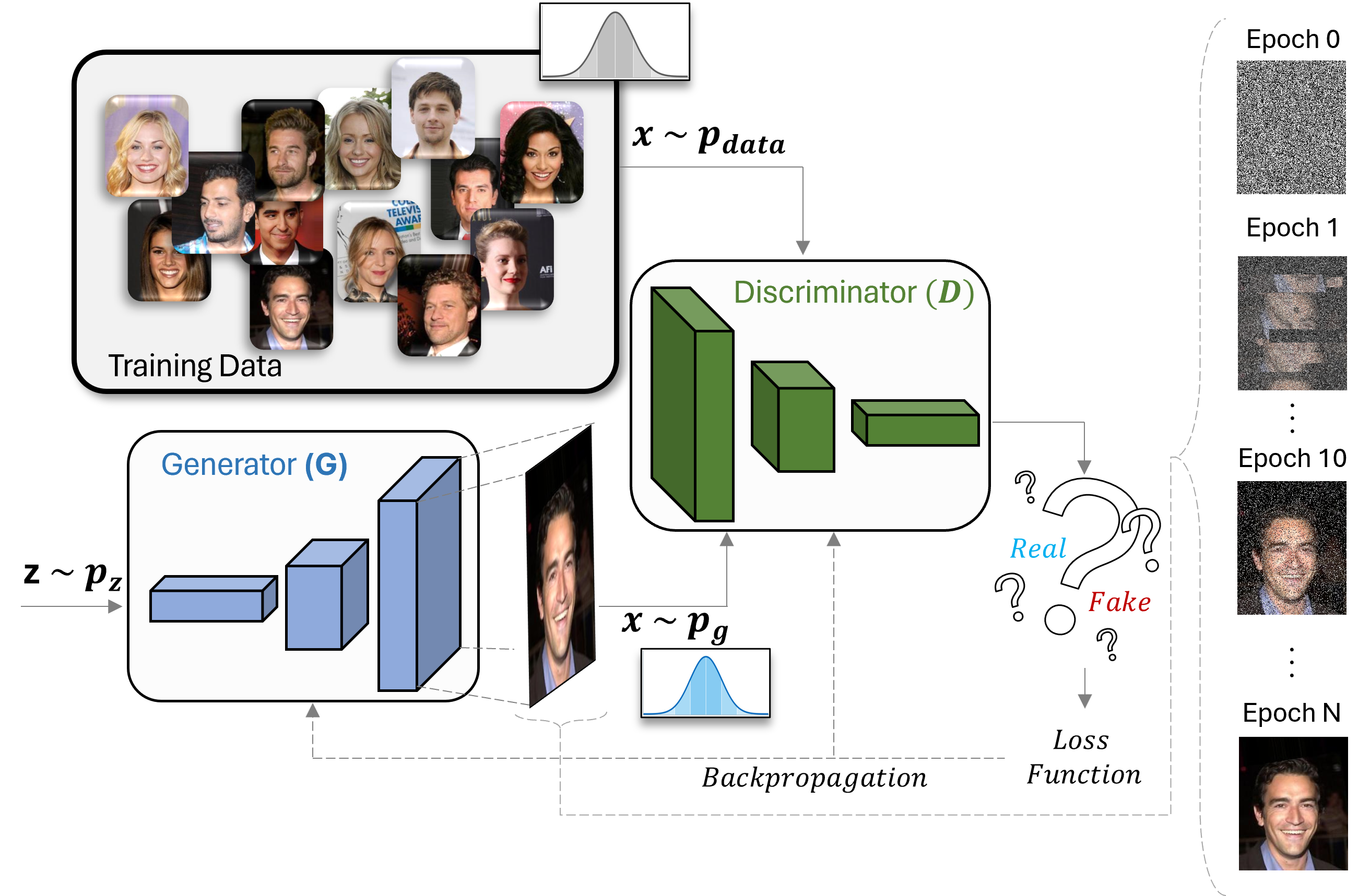}
    \caption{A standard GAN framework. A Generator (\textit{G}) creates data samples from noise, aiming to mimic the training set. A Discriminator (\textit{D}) differentiates between real and \textit{G}-generated data. Training ends when \textit{D} can't distinguish \textit{G}'s images from training samples.}
    \label{fig:gan}
\end{figure}

\begin{figure}[t!]
    \centering
    \includegraphics[width=.8\linewidth]{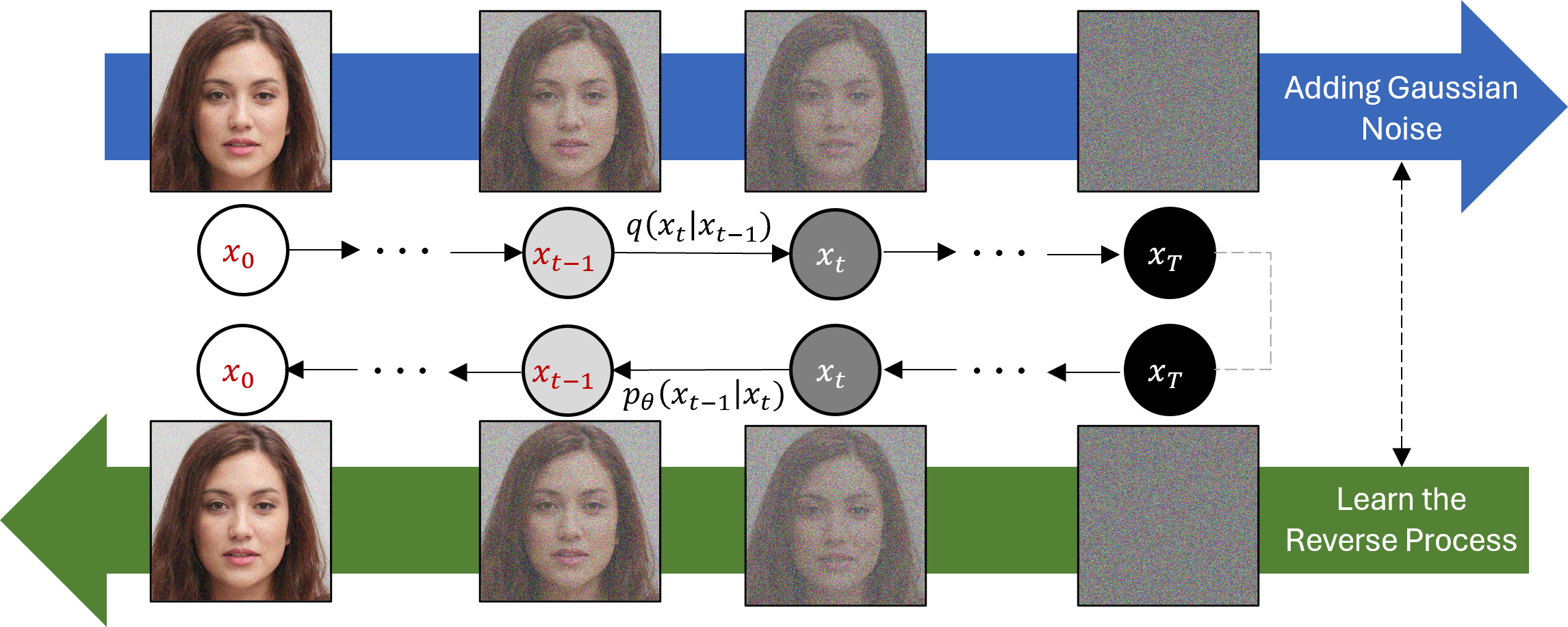}
    \caption{A Diffusion Model architecture. Training data is corrupted with added Gaussian noise. From this data ($x_T$ step), a reverse process is constructed to generate new samples resembling the original ones.}
    \label{fig:dm}
\end{figure}

The emergence of these sophisticated AI technologies has brought about fresh challenges in this domain. One such challenge is the detection of deepfakes, the research area of which is constantly expanding (as shown in Figure~\ref{fig:paperstatistics}). Deepfakes are synthetic media created through generative models based mainly on Generative Adversarial Networks (GANs) and Diffusion Models (DMs)~\citep{goodfellow2014generative}. GANs are composed of a Generator (\textit{G}) and a Discriminator (\textit{D}) trained simultaneously through a competitive process. The Generator is trained to capture the data distribution of the training set \textit{Ts}. The Discriminator is trained to distinguish the images created by \textit{G} from the set \textit{Ts}. When \textit{G} creates images with the same data distribution as \textit{Ts}, \textit{D} will no longer be able to solve its task and the training phase can be considered completed. Currently, researchers demonstrated that synthetic images created by DMs are better than those generated by GAN engines in terms of photorealism, as the creation process follows a more accurate and “controlled” flow. The basic idea of DMs is to iteratively add noise to an input random noise vector for synthetic data generation in order to model complex data distributions. Figure~\ref{fig:gan} and Figure~\ref{fig:dm} show generic GAN and DM schemes related to the creation of synthetic people's faces.

Deepfakes can pose significant challenges in distinguishing real images from manipulated ones, thereby complicating the task of digital forensic investigators: in fact, the problem of deepfake detection has been addressed extensively by the scientific community~\citep{masood2023deepfakes,verdoliva2020media,lin2024detecting}. In this context, preventing cognitive biases in digital forensics becomes crucial to ensure the objectivity and neutrality of judgments.

\section{The Impostor Bias: How AI Media Triggers Bias and Doubt in Perception}
\label{sec:impostorbias}

\begin{figure}[t!]
    \centering
    \includegraphics[width=1\linewidth]{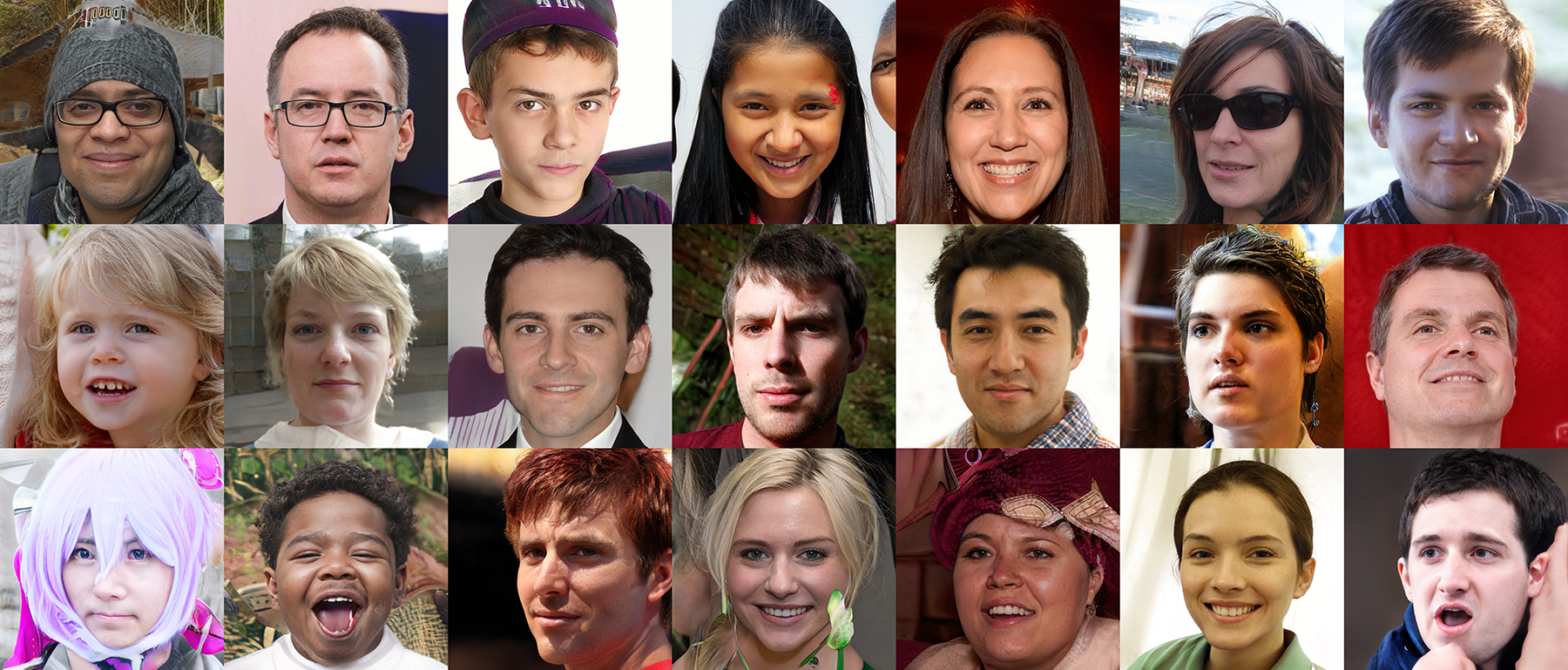}
    \caption{Figures \ref{fig:gan} and \ref{fig:dm} illustrate the architectures of generative models that produce these images. The verisimilitude of these faces could potentially lead observers to question the existence of the depicted individuals, thereby giving rise to the Impostor Bias. Credit:~\citet{guarnera2022ontheExploitation}}
    \label{fig:faces}
\end{figure}

The emergence of GenAI has brought about a sea change in the multimedia landscape, opening up novel avenues for crafting and modifying content. It is largely attributed to the development of a class of machine learning models known as foundation models (FMs)~\citep{rabowsky2023}. These models, which include the likes of ChatGPT released in November 2022, marked the beginning of a new era in Artificial Intelligence. Foundation models are distinguished by their powerful applications to GenAI, which involves the use of models to generate new content and transform existing content. GenAI models can produce high-quality artistic media for visual arts, concept art, music, fiction, literature, video, and animation; distinguishing the real from the fake is becoming increasingly complex, as in the case of human face recognition and its veracity (Figure~\ref{fig:faces}). 

The existence of GenAI and its knowledge can lead to the development of a new type of cognitive bias, which we have identified as the “Impostor Bias”. This is a hypothetical bias with no empirical basis yet, and the term “Impostor” is derived from the “Impostor Syndrome”, a psychological phenomenon in which people doubt their competence and fear being exposed as fraudulent. This bias, however, refers to a different context: distrust of multimedia content generated by Artificial Intelligence. In the context of AI, “Impostor Bias” refers to the tendency to doubt the veracity of multimedia elements such as videos, photos and audios, due to the knowledge that these can be realistically generated by AI models. This bias manifests itself as an a priori distrust, regardless of the quality or context of the multimedia content. The name “Impostor Bias” is appropriate because, just as in Impostor Syndrome, there is a persistent doubt about the veracity of something - in this case, AI-generated media content. Although they may appear authentic and realistic, awareness of the possibility that they are AI-generated “impostors” can lead to doubt and distrust.

The term “bias” is used to describe Impostor Bias because it refers to a systematic and predictable tendency in the way people perceive media content, regardless of the evidence presented. For instance, this is not simply a variation in evaluations due to different levels of knowledge of the evaluator. In other words, even when presented with AI-generated media content that is indistinguishable from the real thing, people with Impostor Bias may still doubt its authenticity due to the knowledge that such content may be AI-generated. This is a bias because it is based on an a priori assumption rather than an objective assessment of the content itself. Furthermore, the term “bias” implies that this tendency can lead to distortions in perception and judgement. For example, “Impostor Bias” may lead people to discard or devalue authentic media content because they perceive it as potentially false or misleading.

\begin{figure}[b!]
    \centering
    \includegraphics[width=.8\linewidth]{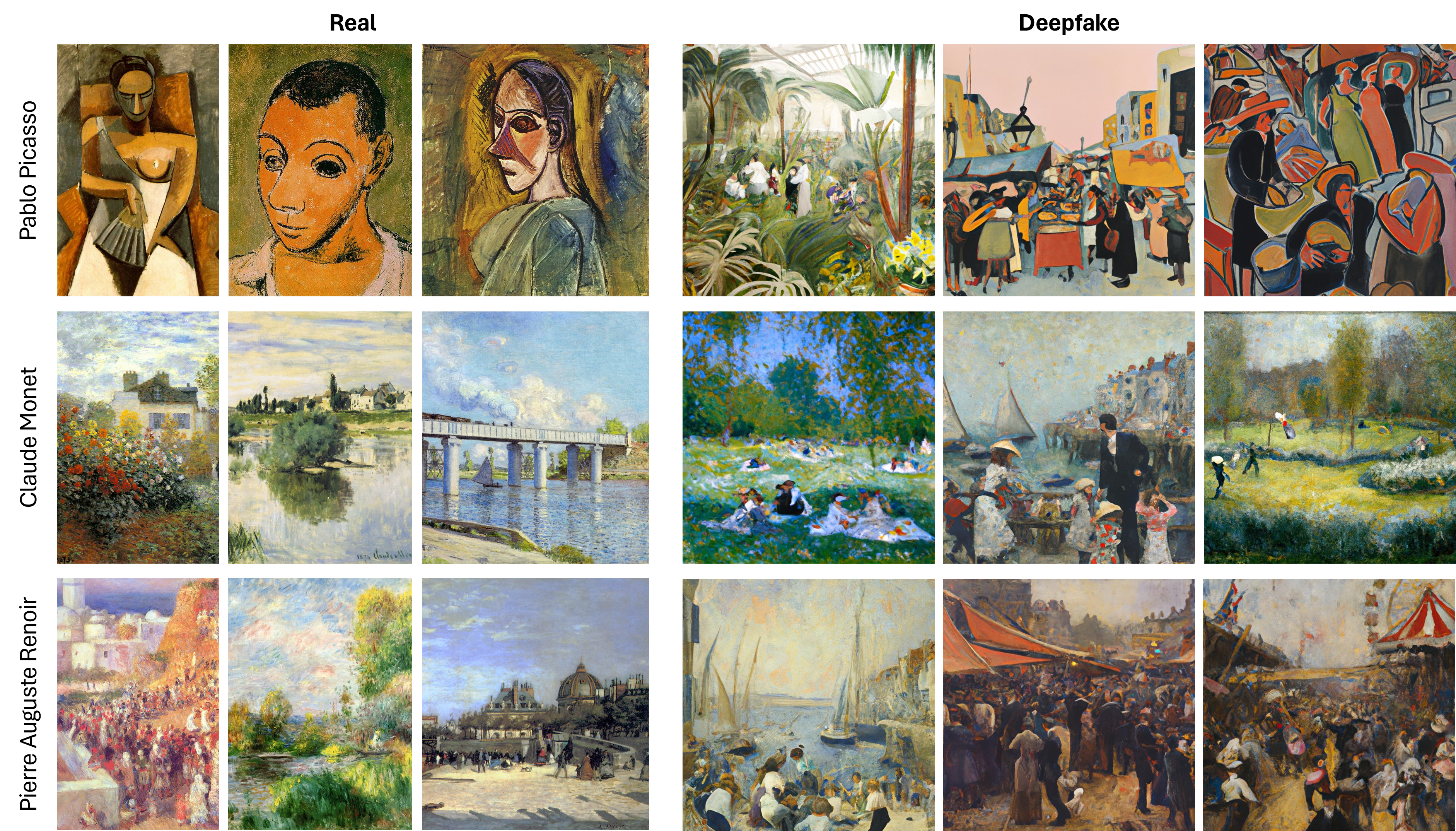}
    \caption{Real vs deepfake images of famous artists.}
    \label{fig:artist}
\end{figure}

Judicial practitioners and investigators may struggle with determining the authenticity of multimedia content, especially as AI-generated media becomes more realistic and descriptive. The risk of falling into cognitive traps or suspecting original content as AI-generated underscores the complexity of this issue. Additionally, GenAI raises concerns about artwork counterfeiting, copyright infringement, and the potential for forged masterpieces to be sold as genuine.  Indeed, such technologies are now capable of simulating the technique and artistic style of the most famous artists, thereby compromising the ability to correctly discern between real and simulated works~\citep{epstein2023art,leotta2023not} (Figure~\ref{fig:artist}). Forensic examiners from 21 countries showed limited understanding and appreciation of cognitive bias, with fewer than half supporting blind testing, highlighting the need for procedural reforms to blind them to potentially biasing information~\citep{Kukucka2017Cognitive}.

\section{Generative AI and Deepfake Detection Methods}
\label{sub:detection}

The mitigation of the potential impacts of the hypothetical “Impostor Bias" could significantly rely on Generative AI and deepfake detection methods. As AI-generated media products become increasingly realistic \citep{verdoliva2020media,de-Lima-Santos2021Artificial}, the prevalence of “Impostor Bias" is expected to rise, posing significant challenges in various sectors, including the justice system. In this context, technologies for deepfake detection become increasingly important. These technologies are evolving to not only recognize manipulated or altered multimedia artifacts but also those generated from scratch using descriptive textual inputs.

Detecting content generated by Generative AI, such as ChatGPT, is a rapidly evolving field of study. Recent research has focused on developing machine learning tools capable of distinguishing between human-written text and machine-generated content \citep{Prillaman_2023,Editorial_2023,Weber_Wulff_2023,perkins2024genai,liu2024detectability}. These tools, such as the one described in a study by \citet{perkins2024genai}, analyze various features of writing style, including variation in sentence lengths, and the frequency of certain words and punctuation marks. However, the efficacy of these detection tools can be significantly reduced when they are confronted with machine-generated content that has been modified using techniques designed to evade detection \citep{perkins2024genai}. For instance, the study found that the detectors' already low accuracy rates (39.5\%) showed major reductions in accuracy (17.4\%) when faced with manipulated content. Despite these challenges, the development of more accurate and robust GenAI detection methods continues to be a critical area of research, given the increasing prevalence of AI-generated content in various domains \citep{perkins2024genai,liu2024detectability}. For example, a tool called CheckGPT has been developed, which examines 20 features of writing style to determine whether an academic scientist or ChatGPT wrote a piece of text \citep{liu2024detectability}. The tool was found to be highly accurate, achieving an average classification accuracy of 98\% to 99\% for task-specific discipline-specific detectors and the unified detectors.

\begin{figure}[t!]
    \centering
    \includegraphics[width=1\linewidth]{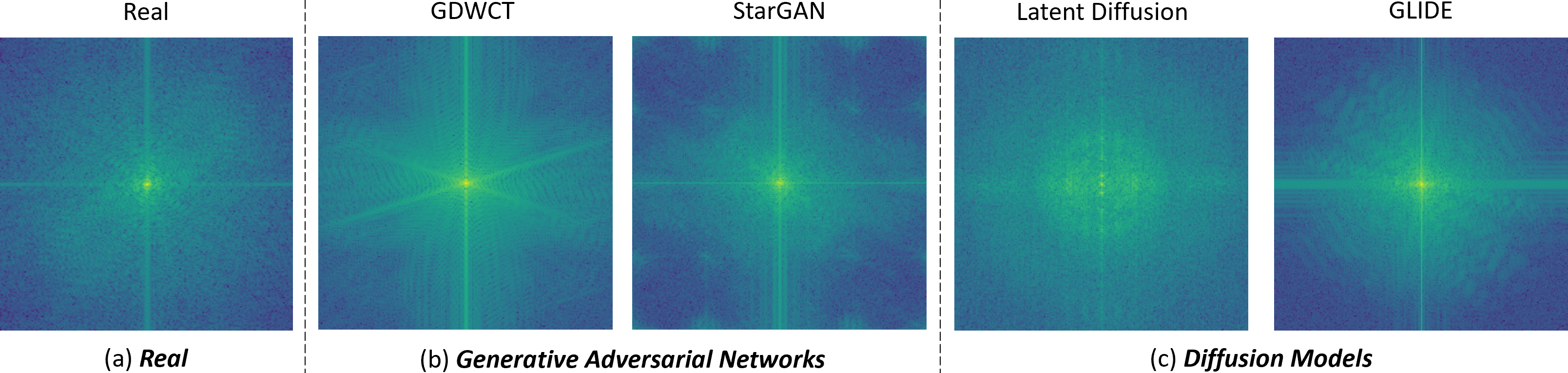}
    \caption{Fourier spectrum of different categories of data: (a) real images; (b) images generated by two GAN architectures (GDWCT \citep{cho2019image}  and StarGAN~\citep{choi2018stargan}); (c) images generated by two DM architectures (Latent Diffusion \citep{ramesh2022hierarchical} and GLIDE \citep{nichol2022glide}). The abnormal frequencies (light peaks) are mainly visible in the images generated by artificial intelligence engines.}
    \label{fig:fourier}
\end{figure}

Furthermore, researchers have demonstrated that generative engines leave traces on synthetic content that can be identified and detected in the frequency domain~\citep{guarnera2020preliminary,zhang2019detecting,Marra2019DoGL,Giudice_JI_2021, Dzanic_NEURIPS_2020,Durall_CVPR_2020} (Figure~\ref{fig:fourier}). These traces are characterized by both the network architecture (number and type of layers) and its specific parameters~\citep{yu2019attributing}.  
In order to distinguish real data from deepfake,~\citet{guarnera2020Deepfake,guarnera2020fighting} proposed methods based on the Expectation-Maximization~\citep{moon1996expectation} algorithm capable of capturing traces defined as the correlation of pixels left by convolutional layers.


~\citet{wang2020cnn} used ResNET-50 to distinguish real and ProGAN-generated images, showing generalization across different GANs. FakeSpotter, proposed by~\citet{wang2021fakespotter}, detects GAN-generated faces by monitoring CNN neuron behaviors. Vision Transformer-based solutions for deepfake detection have also been proposed~\citep{wodajo2021Deepfake,coccomini2022combining,heo2023deepfake,wang2022m2tr}.~\citet{wodajo2021Deepfake} combined transformers with a convolutional network to extract patches from detected faces in videos. Several studies~\citep{yu2019attributing,yu2021artificial,girish2021towards,asnani2023reverse,yu2020responsible,guarnera2022ontheExploitation,guarnera2024mastering} have explored identifying specific GAN models used in creation (Model Attribution Task).~\citet{guarnera2022ontheExploitation} distinguished 100 StyleGAN2 instances using ResNET-18~\citep{he2016deep} and metric learning~\citep{liu2012metric}, demonstrating the method's effectiveness in deepfake model recognition.

The scientific community is also working extensively on the creation of advanced techniques for the detection of synthetic images created by diffusion models.~\citet{corvi2023detection} have been trying to understand how difficult it is to distinguish synthetic images generated from diffusion models from real ones, and whether current state-of-the-art detectors are suitable for this task.~\citet{sha2022fake} proposed DE-FAKE, a machine-learning classifier-based method for diffusion model detection on four popular text-image architectures.~\citet{guarnera2023level} proposed a hierarchical approach based on different architectures in order to define: whether the image is real or manipulated via any generative architecture (AI-generated); the specific framework used among GAN or DM; defines the specific generative architecture used among a predefined set. Furthermore, another practical digital forensic tool, Transfer learning-based Autoencoder with Residuals (TAR), was proposed~\citep{lee2021tar}. The ultimate goal of TAR was to develop a unified model to detect various types of deepfake videos with high accuracy, with only a small number of training samples that can work well in real-world settings. A short summary of these methods can be found in Table~\ref{tab:methods}.

\begin{table}[]
\centering
\resizebox{\textwidth}{!}{%
\begin{tabular}{@{}llll@{}}
\toprule
\textbf{Reference} &
  \textbf{Generation Models} &
  \textbf{Database(s) Used} &
  \textbf{Precision (avg)} \\ \midrule
\citet{he2016deep} &
  StyleGAN, StyleGAN2-ADA &
  FFHQ (Flickr-Faces-HQ) &
  96.2\% \\
\citet{wang2020cnn} &
  ProGAN &
  CelebA &
  99.1\% \\
\citet{guarnera2023level} &
  \begin{tabular}[c]{@{}l@{}}GANs: (AttGAN, CycleGAN, GDWCT, IMLE, ProGAN,\\ StarGAN, StarGAN-v2, StyleGAN, StyleGAN2)\\ DMs: (DALL·E 2, GLIDE, Latent Diffusion, Stable Diffusion)\end{tabular} &
  CelebA, FFHQ, ImageNet &
  \begin{tabular}[c]{@{}l@{}}97,6\% (Level 1)\\ 98,0\% (Level 2)\\ 97,8\% (Level 3, GANs)\\ 98,0\% (Level 3, DMs)\end{tabular} \\
\citet{wang2021fakespotter} &
  StyleGAN, StyleGAN2, BigGAN, ProGAN &
  FaceForensics++ &
  90.6\% \\
\citet{wodajo2021Deepfake} &
  \begin{tabular}[c]{@{}l@{}}FaceSwap, Face2Face, FaceShifter, NeuralTextures,\\ DeepFakeDetection\end{tabular} &
  FaceForensics++, UADFV &
  91.5\% \\
\citet{lee2021tar} &
  FaceSwap, Face2Face, DeepFake, NeuralTextures &
  FaceForensics++ &
  \begin{tabular}[c]{@{}l@{}}98.0\% deepfake type detection\\ 89.5\% on DW videos\end{tabular} \\
\citet{sha2022fake} &
  GLIDE, Latent Diffusion, Stable Diffusion, DALL·E 2 &
  MSCOCO (a), Flickr30k (b) &
  90.2\%(a), 84.6\% (b) \\ \bottomrule
\end{tabular}%
}
\caption{A summary of the discussed deepfake detection methods.}
\label{tab:methods}
\end{table}

Experimental results of all these methods show that, in general, all generative models leave unique traces that can solve all previously listed tasks with high accuracy. Therefore, these methods can be used in order to help the general user in countering what we called the \textit{Impostor Bias} phenomenon.
However, we want to highlight an important element of the previously listed methods and not just deepfake detection. All methods in the literature achieve extremely high results in “constrained"  contexts, that is, with architectures known a priori. In practice, current deepfake detectors fail to generalize with synthetic images generated by novel architectures (different from those used during the training procedure), resulting in a drastic drop in classification performance. Some recent new methods~\citep{dong2023implicit,coccomini2023generalization} published by the scientific community seem to be good starting points in order to achieve generalization.

\section{Discussion}
\label{sec:discussion}

The concept of Impostor Bias, though not yet widely recognized, is increasingly relevant in the era of deepfakes and digital forensics. The recent Ukraine war, marked by propaganda and misinformation on social media \citep{ciuriak_social_2022,suciu_is_2022}, has heightened this bias. The constant exposure to videos, photos, and statements fosters an inherent suspicion of media manipulation or AI generation \citep{linehan_deepfakes_2023}. This "digital warfare" has led to immediate doubt and scrutiny of all published media, sometimes uncovering deepfakes \citep{bond_how_2023}. The prevalence of Impostor Bias will persist, especially with sensitive content like wartime communications \citep{linehan_deepfakes_2023}. Thus, novel techniques and strategies are essential for effective mitigation.

The unique nature of the Impostor Bias necessitates specialized approaches beyond the strategies discussed in Section 4. While those strategies offer valuable insights, addressing Impostor Bias effectively requires a combination of targeted measures. This includes shielding analysts from irrelevant information and providing specific training on cognitive bias risks. Additionally, proficiency testing and regular assessments maintain practitioner competence. However, to truly counter Impostor Bias, advanced technical tools for synthetic content detection are essential. Practical deep learning algorithms can detect generative models' traces, aiding investigators in distinguishing authentic evidence. Given the success of deepfake detection algorithms \citep{pei2024deepfake,electronics13030585}, these tools can assist in counteracting Impostor Bias and other digital biases. Similar detection methods for forgery images \citep{baumy2022,Zanardelli2023,Singh2024} and multimedia data manipulation \citep{Galante2023,DUNSIN2024301675} further reinforce the value of algorithmic assistance in avoiding bias. The increasing sophistication of digital data demands the use of advanced deepfake detection techniques to maintain trust and accuracy in forensic examinations.

Detection methods play a crucial role in identifying deepfakes, but interpreting and explaining the results is equally important. While some methods use classic machine learning techniques, others employ deep learning techniques, each with its advantages and limitations. Classic machine learning approaches are more interpretable but lack robustness, while deep learning methods offer higher performance and robustness but are more complex to explain. The question of trusting the predictions of detection algorithms is indeed a separate but related study, as explored by \citet{Mathews_Trivedi_House_Povolny_Fralick_2023}, \citet{app12083926}, and others \citep{canteroarjona2024deepfake,pinhasov2024xaibased,Pandey_Singh_Malik_Kumar_2024,pontorno2024exploitation}. These studies enhance the “trust" in detection methods and underscore the need for further exploration of cognitive biases in digital forensic science. Creating specialized training programs for practitioners can enhance awareness, reduce the influence of biases, and preserve the objectivity of judgments by disregarding irrelevant elements and separating personal experiences.

\section{Conclusions and Future Works}
\label{sec:conclusions}

In this article, we have explored the significant impact of cognitive biases in the fields of forensics and digital forensics, specifically focusing on confirmation bias, anchoring bias, and hindsight bias. To reduce the influence of these biases, we have discussed strategies such as game-based interventions and the Linear Sequential Unmasking-Expanded approach. With the growing prevalence of deepfakes, synthetic media that can manipulate or impersonate individuals, the integrity of digital evidence is at risk. We have surveyed current deepfake detection methods and their limitations, introducing the novel concept of Impostor Bias. This bias, influenced by the widespread use of deepfakes, may lead to false negatives and reduced confidence in digital forensic findings. To address these challenges, we propose future research directions, including the development of advanced deepfake detection methods and a deeper understanding of the factors contributing to Impostor Bias. Furthermore, we emphasize the importance of interventions to mitigate this bias and the need to explore the ethical, legal, and social implications of deepfakes. By doing so, we aim to enhance the reliability and integrity of digital forensic practices, ensuring the accuracy and objectivity of forensic investigations.

\section{Acknowledgments}
\label{sec:ackn}
This research is supported by Azione IV.4 - ``Dottorati e contratti di ricerca su  tematiche dell’innovazione" del nuovo Asse IV del PON Ricerca e Innovazione 2014-2020 “Istruzione e ricerca  per il recupero - REACT-EU”- CUP: E65F21002580005.


\bibliographystyle{elsarticle-harv} 
\bibliography{cas-refs}





\end{document}